# Online Contrastive Divergence with Generative Replay: Experience Replay without Storing Data

**Decebal Constantin Mocanu · Maria Torres Vega ·
Eric Eaton · Peter Stone · Antonio Liotta**

**Abstract** Conceived in the early 1990s, Experience Replay (ER) has been shown to be a successful mechanism to allow online learning algorithms to reuse past experiences. Traditionally, ER can be applied to all machine learning paradigms (i.e., unsupervised, supervised, and reinforcement learning). Recently, ER has contributed to improving the performance of deep reinforcement learning. Yet, its application to many practical settings is still limited by the memory requirements of ER, necessary to explicitly store previous observations. To remedy this issue, we explore a novel approach, Online Contrastive Divergence with Generative Replay (OCD$_{GR}$), which uses the generative capability of Restricted Boltzmann Machines (RBMs) instead of recorded past experiences. The RBM is trained online, and does not require the system to store any of the observed data points. We compare OCD$_{GR}$ to ER on 9 real-world datasets, considering a worst-case scenario (data points arriving in sorted order) as well as a more realistic one (sequential random-order data points). Our results show that in 64.28% of the cases OCD$_{GR}$ outperforms ER and in the remaining 35.72% it has an almost equal performance, while having a considerably reduced space complexity (i.e., memory usage) at a comparable time complexity.

**Keywords** generative replay · experience replay · unsupervised learning · online learning · density estimation · restricted Boltzmann machines · deep learning

Decebal Constantin Mocanu
Department of Electrical Engineering, Eindhoven University of Technology, Eindhoven, the Netherlands
Tel.: +31 40-247 5394
E-mail: d.c.mocanu@tue.nl

Maria Torres Vega
Department of Electrical Engineering, Eindhoven University of Technology, Eindhoven, the Netherlands

Eric Eaton
Department of Computer and Information Science, University of Pennsylvania, Philadelphia, USA

Peter Stone
Department of Computer Science, The University of Texas at Austin, Austin, USA

Antonio Liotta
Department of Electrical Engineering, Eindhoven University of Technology, Eindhoven, the Netherlands

**1 Introduction**

Experience Replay (ER) (Lin, 1992) (dubbed interleaved learning in McClelland et al (1995)) has been shown to be a successful mechanism in helping online learning algorithms to reuse past experiences. In ER, the data acquired during the online learning process is stored explicitly and presented repeatedly to the online learning algorithm, such as reinforcement learning (RL) (Adam et al, 2012), deep reinforcement learning (DRL) (Mnih et al, 2015), or supervised learning (McClelland et al, 1995). The ER process enables the learner to achieve good performance from limited training data, and helps to break temporal correlations in the observations which go against the i.i.d assumptions of many stochastic gradient-based algorithms (Schaul et al, 2015). Since ER uses recorded data in chunks, it has sometimes been deemed a batch learning approach (Kalyanakrishnan and Stone, 2007). In general, ER focuses on the reuse of observed data in its raw form as stored in memory, replaying it to the online learner. However, this causes ER to scale poorly, as the memory requirements increase as the environment and system requirements increase. One common practice is to limit the available memory of the ER mechanism and to either 1.) discard the oldest experiences as the memory buffer becomes full and/or 2.) prioritize the experiences (Schaul et al, 2015).

From a biological sense of memory (i.e., hippocampal replay in McClelland et al (1995)), the human brain does *not* store all observations explicitly, but instead it *dynamically generates approximate reconstructions* of those experiences for recall. This idea has also been applied to online learning through model-based learning as an alternative to ER. Such approaches indirectly reuse experiences by first modeling the environment, and then using that model to generate new data. This procedure is used by Dyna and other model-based learning approaches (Sutton et al, 2008). Building a model will generally require less memory than storing the raw data, and can diminish the effects of noise in the observations. However, model learning incurs additional computational costs and, more importantly, will introduce modeling errors that can significantly decrease performance (Sutton et al, 2008). For this reason, it is necessary to look for alternatives that are able to scale effectively (which is one of the biggest issues in ER) and yield performance results that are comparable with those obtained under ER, without increased computational complexity.

At the same time, Restricted Boltzmann Machines (RBMs) (Smolensky, 1987), the original building blocks in deep learning models (Bengio, 2009), besides providing in an unsupervised manner good weights for the deep belief networks initialization (LeCun et al, 2015), have been shown to be very good density estimators and to have powerful generative capabilities (Salakhutdinov and Murray, 2008; Mocanu et al, 2016). Due to these capabilities RBMs and models derived from them have been successfully applied to various problems also as standalone models. Examples of these applications are: modeling human choice (Osogami and Otsuka, 2014), collaborative filtering (Salakhutdinov et al, 2007), information retrieval (Gehler et al, 2006), transfer learning (Ammar et al, 2013), or multi-class classification (Larochelle and Bengio, 2008). However, in all of the above settings RBMs have been used offline using offline training algorithms. This reduces drastically their capabilities to tackle real-world problems which can not be handled on server clouds using GPU computing, and require fast training algorithms capable of continuous learning when the the environment is changing. For example, in the world of wireless sensor networks which is by definition an environment with low-resources (e.g., memory, computational power, low energy) devices to perform anomaly detection in time series directly in the wireless nodes would help improving the network capacity on various components (e.g., lifetime, avoid data traffic congestions), as exemplified in Bosman et al (2017). We then hypothesize that due to their density estimation capabilities RBMs which have been used recently to estimate the similarity between data distributions in various domains (e.g., image quality assessment (Mocanu et al, 2014), Markov decision processes (Ammar et al, 2014)), could be used to perform anomaly detection directly on any wireless node if they would have available an online training algorithm which has low memory requirements (best case is to store none of the historical data). Still, to our knowledge, there are no dedicated algorithms to train RBMs in a fully online manner—the only currently available solution is to employ ER mechanisms with memory, by following the successful examples from other deep learning models (e.g., Mnih et al (2015)).

In this paper, we combine the generative capabilities of RBMs with the biological inspiration behind experience replay, yielding a novel algorithm to train RBMs in online settings, which we call *Online Contrastive Divergence with Generative Replay* (OCD$_{GR}$). In comparison with state-of-the-art ER techniques, OCD$_{GR}$ acts more like the experience replay concept in a biological sense. Instead of explicitly storing past observations in memory, it generates new training data dynamically to represent historical observations, using the generative capabilities of the RBM itself. In contrast to model-based learning approaches, which learn models for the environment (Sutton et al, 2008), OCD$_{GR}$ relies on the underlying RBM, which models only the observed data distribution—a substantially easier problem. OCD$_{GR}$ derived methods may have a wide applicability to a variety of tasks (e.g., regression, classification, reinforcement learning, anomaly detection), but in this paper we focus on demonstrating the benefits of OCD$_{GR}$ over current ER approaches on the RBMs main task (i.e., distribution estimation), this being a must have for any further developments. Thus, using 9 real-world datasets we show how OCD$_{GR}$ outperforms ER in training RBMs, while having reduced memory requirements and an equivalent time complexity.

The remainder of this paper is organized as follows. Section 2 presents background knowledge about experience replay and restricted Boltzmann machines for the benefit of the non-specialist reader. Section 3 introduces our proposed method, while Section 4 describes the experiments performed and assesses the results. Finally, Section 5 concludes the paper and presents further research directions.

## 2 Background and Related Work

In this section, we first discuss related work on ER. Next, background informations on RBMs and their offline training methods are presented.

### 2.1 Experience Replay (ER)

Experience replay was first introduced for RL in Lin (1992) and for supervised learning in McClelland et al (1995). Later on, a number of methods have been proposed for ER, aiming to model the environment and to optimize the performance of online learning. A complete review of ER applicability does not constitute a goal of this paper, but as an example, Kalyanakrishnan and Stone (2007) showed that the standard RL and the batch approaches (ER) are easily comparable in terms of performance. Recently, ER and its variants have contributed to improving deep reinforcement learning (DRL) (Mnih et al, 2015). Narasimhan et al (2015) proposed a form of re-sampling in the context of DRL, separating the learner experience into two parts: one for the positive rewards and one for the negative. Ororbia et al (2015) proposed an online semi-supervised learning algorithm using deep hybrid Boltzmann machines and denoising autoencoders. However, all of these approaches require memory to explicitly store past observations for recall, making them less suitable to the online learning setting.

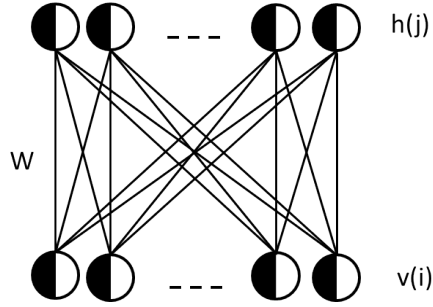

Fig. 1: Restricted Boltzmann Machine architecture.

## 2.2 Restricted Boltzmann Machines

In this paper, we use the generative capabilities of Restricted Boltzmann Machines (RBMs) to dynamically generate new training data during the online learning process, instead of explicitly storing and recalling past observations. We next review the mathematical details of RBMs. They were introduced by Smolensky (1987) as a powerful model to learn a probability distribution over its inputs. Formally, RBMs are generative stochastic neural networks with two binary layers: the hidden layer $\mathbf{h} = [h_1, h_2, .., h_{n_h}] \in \{0, 1\}^{n_h}$, and the visible layer $\mathbf{v} = [v_1, v_2, .., v_{n_v}] \in \{0, 1\}^{n_v}$, where $n_h$ and $n_v$ are the numbers of hidden neurons and visible neurons, respectively. In comparison with the original Boltzmann machine (Ackley et al, 1985), the RBM architecture (Figure 1) is restricted to be a complete bipartite graph between the hidden and visible layers, disallowing intra-layer connections between the units. The energy function of an RBM for any state $\{\mathbf{v}, \mathbf{h}\}$ is computed by summing over all possible interactions between neurons, weights, and biases as follows:

$$E(\mathbf{v}, \mathbf{h}) = -\mathbf{a}^\mathsf{T}\mathbf{v} - \mathbf{b}^\mathsf{T}\mathbf{h} - \mathbf{h}^\mathsf{T}\mathbf{W}\mathbf{v} \ , \qquad (1)$$

where $\mathbf{W} \in \mathbb{R}^{n_h \times n_v}$ is the weighted adjacency matrix for the bipartite connections between the visible and hidden layers, and $\mathbf{a} \in \mathbb{R}^{n_v}$ and $\mathbf{b} \in \mathbb{R}^{n_h}$ are vectors containing the biases for the visible and hidden neurons, respectively. For convenience, we can bundle the RBM's free parameters together into $\Theta = \{\mathbf{W}, \mathbf{a}, \mathbf{b}\}$. Functionally, the visible layer encodes the data, while the hidden layer increases the learning capacity of the RBM model by enlarging the class of distributions that can be represented to an arbitrary complexity (Taylor et al, 2011). Due the binary state of the neurons, the free energy of the visible units may be computed as (Bengio, 2009):

$$\mathcal{F}(\mathbf{v}) = -\mathbf{a}^\mathsf{T}\mathbf{v} - \sum_j \log(1 + \exp(b_j + \mathbf{W}_{j:}\mathbf{v})) \ , \qquad (2)$$

where $\mathbf{W}_{j:}$ represents the $j^{th}$ row of the matrix $\mathbf{W}$. The activations of the hidden or visible layers are generated by sampling from a sigmoid $\mathcal{S}(\cdot)$ according to: $P(\mathbf{h} = 1|\mathbf{v}, \Theta) = \mathcal{S}(\mathbf{b} + \mathbf{W}\mathbf{v})$ and $P(\mathbf{v} = 1|\mathbf{h}, \Theta) = \mathcal{S}(\mathbf{a} + \mathbf{W}^\mathsf{T}\mathbf{h})$ .

2.3 Offline RBM Training via Contrastive Divergence

The RBM parameters can be learned effectively by following the log-likelihood gradient computed over a training set $\mathcal{D}$, with $n_v$-dimensional binary instances. The log-likelihood gradient is given by:

$$\mathbb{E}_{\hat{P}}\left[\frac{\partial(logP(\mathbf{v}))}{\partial \theta}\right] = -\mathbb{E}_{\hat{P}}\left[\frac{\partial \mathcal{F}(\mathbf{v})}{\partial \theta}\right] + \mathbb{E}_{P}\left[\frac{\partial \mathcal{F}(\mathbf{v})}{\partial \theta}\right] \;, \qquad (3)$$

where $\hat{P}$ represents the empirical distribution of $\mathcal{D}$ and $\mathbb{E}_P$ is the expectation computed under the model distribution (Bengio, 2009). However, sampling from $P$ to compute the free energy and running long Monte-Carlo Markov Chains (MCMC) to obtain an estimator of the log-likelihood gradient is usually intractable. Due to this intractability, Hinton (2002) proposed an approximation method called Contrastive Divergence (CD), which solves the above problem by making two approximations. The first approximation is to replace the average over all possible inputs from the second term of Equation 3 by a single sample. The second approximation is to run each MCMC chain for only a specific number of steps ($n_{CD}$), starting from a data point $\mathbf{v}^0 \in \mathcal{D}$, as follows:

$$\mathbf{v}^0 \xmapsto{P(\mathbf{h}|\mathbf{v}^0)} \mathbf{h}^0 \xmapsto{P(\mathbf{v}|\mathbf{h}^0)} \mathbf{v}^1 \xmapsto{P(\mathbf{h}|\mathbf{v}^1)} \mathbf{h}^1 \dashrightarrow \mathbf{v}^{n_{CD}} \longmapsto \mathbf{h}^{n_{CD}} \;.$$

The free parameters can then be updated afterwards via:

$$\Delta \Theta = \frac{\partial \mathcal{F}(\mathbf{v}^0)}{\partial \theta} - \frac{\partial \mathcal{F}(\mathbf{v}^{n_{CD}})}{\partial \theta} \;, \qquad (4)$$

yielding the following update rules for the free parameters of binary RBMs:

$$\begin{aligned}
\Delta W_{ji} &\propto v_i^0 h_j^0 - v_i^{n_{CD}} h_j^{n_{CD}} &&\text{for } 1 \leq i \leq n_v, 1 \leq j \leq n_h \\
\Delta a_i &\propto v_i^0 - v_i^{n_{CD}} &&\text{for } 1 \leq i \leq n_v \\
\Delta b_j &\propto h_j^0 - h_j^{n_{CD}} &&\text{for } 1 \leq j \leq n_h \;.
\end{aligned} \qquad (5)$$

Several other variants of contrastive divergence have been proposed to train RBMs offline. Examples of these are: persistent contrastive divergence (Tieleman, 2008), fast persistent contrastive divergence (Tieleman and Hinton, 2009), parallel tempering (Desjardins et al, 2010), and the replace of the Gibbs sampling with a transition operator to obtain a faster mixing rate and an improved learning accuracy without affecting the computational costs (Brügge et al, 2013). Yet, in this paper we use the original CD (Hinton, 2002), as it is easily adaptable to online settings, and at the same time it is widely used and allows for a direct comparison with other results reported in the literature.

## 3 Online Contrastive Divergence with Generative Replay

This section presents our novel algorithm to train RBMs online: *Online Contrastive Divergence with Generative Replay* (OCD$_{GR}$). Our approach adapts the standard CD algorithm (see Section 2.3) to the online learning setting, and uses dynamic generation of data as a replay mechanism. We show how an RBM trained via OCD$_{GR}$ can have the same functionality as training via ER. However, OCD$_{GR}$ provides the significant advantage of not needing to explicitly store past observations in memory, substantially reducing its space complexity. To our knowledge, this capability is unique, since other state-of-the-art experience replay mechanisms require a memory dataset to store historical data.

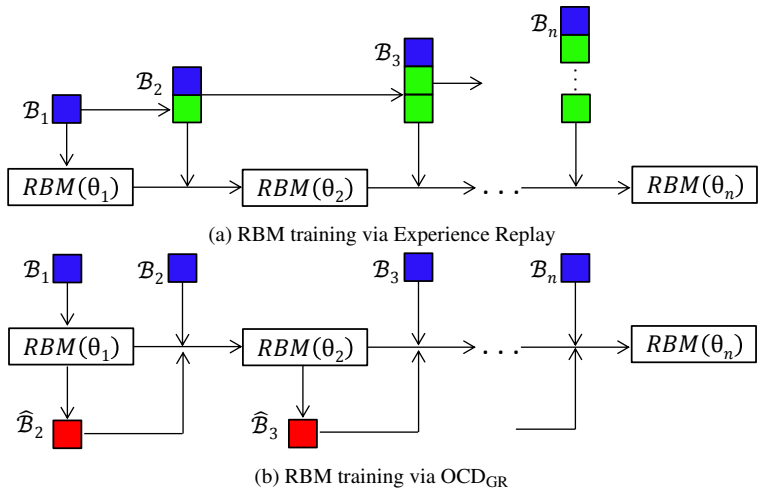

Fig. 2: A comparison of ER with memory and OCD$_{GR}$ for training RBMs online. Each subscript $\cdot_t$ represents a discrete time step $t$. $\mathcal{B}_t$ represents a batch of observed data between $t-1$ and $t$, while $\hat{\mathcal{B}}_t$ represents samples generated by the RBM model using the free parameters $\Theta_{t-1}$ (i.e., the parameters values at time $t-1$).

### 3.1 Intuition and Formalism

Our algorithm is motivated by the fact that hippocampal replay (McClelland et al, 1995) in the human brain does *not* recall previous observations explicitly, but instead it generates *approximate reconstructions* of those past experiences for recall. At the same time, RBMs can generate good samples of the incorporated data distribution via Gibbs sampling (Bengio, 2009). Intuitively, by using those generated samples (instead of previous observations from stored memory as in ER) during the online training process, any RBM model can retain knowledge of past observations while learning new ones.

Before entering into the technical details of our proposed method, we mention that further on we use the following notations: $\mathcal{B}_t$ represents a batch of observed data between time $t-1$ and $t$, while $\hat{\mathcal{B}}_t$ represents samples generated by the RBM model using the free parameters $\Theta_{t-1}$ (i.e., the parameters values at time $t-1$). Figure 2 summarizes the main differences between the OCD$_{GR}$ (Figure 2b) and ER mechanisms (Figure 2a) for training RBMs online, showing how ER explicitly stores previous observations in memory for recall, while OCD$_{GR}$ dynamically generates samples from its current model of the input data distribution. Clearly, the memory used by ER increases linearly with the amount of data observed (up to a fixed limit for memory-bounded ER methods), while by contrast, OCD$_{GR}$ maintains the same memory footprint throughout the training process. Also, note that OCD$_{GR}$ has the Markov property that $P(\Theta_t)$ depends only upon $\Theta_{t-1}$ and $\mathcal{B}_t$, while the ER mechanism with memory does not, since for ER, $P(\Theta_t)$ is dependent upon $\{\Theta_{t-1}, \mathcal{B}_1, \mathcal{B}_2, \ldots, \mathcal{B}_t\}$. This is an important aspect for an algorithm which runs for an indefinite amount of time, as may occur in many real-time systems. Formally, OCD$_{GR}$ is a continuous-time Markov chain with finite (countable) state space $\mathcal{X}$, given by a family $\{RBM_t = RBM(t)\}_{t>0}$ of $\mathcal{X}$ such that:

1. $t \mapsto RBM(t)$ are right-continuous step functions, and

2. $\forall s, s_1, ..., s_k \in \mathcal{X}$, and every sequence of times $t_1 < t_2 < ... < t_k < t_{k+1}$, it holds that:

$$P\big(RBM(t_{k+1}) = s \mid RBM(t_k) = s_k, ..., RBM(t_1) = s_1\big)$$
$$= P\big(RBM(t_{k+1}) = s \mid RBM(t_k) = s_k\big) \ .$$

The second condition is the natural continuous-time analogue of the Markov property, and it requires that the future is conditionally independent of the past given the present RBM. A continuous time Markov chain is a non-lattice semi-Markov model, so it has no concept of periodicity. Consequently, the long-runtime averages equals the limiting probabilities, and it has an equilibrium distribution.

3.2 Algorithm

OCD$_{GR}$ is presented as Algorithm 1. As input, the algorithm accepts various meta-parameters, two of them being specific for OCD$_{GR}$, while the others are common to all RBM models (Algorithm 1, line 2). The two meta-parameters specific for OCD$_{GR}$ are the number of Gibbs sampling steps for the generation of the new training data points ($n_{Gs}$), and the number of new data points generated by the RBM with Gibbs sampling ($n_{\hat{B}}$). The common RBMs meta-parameters include the number of hidden neurons ($n_h$), the number of visible neurons ($n_v$) (which is given by the dimensionality of the data), the number of CD steps ($n_{CD}$), the number of training epochs ($n_E$), the number of data points stored in a mini-batch before the RBM parameters are updated ($n_B$), the learning rate ($\alpha$), the momentum ($\rho$), and the weight decay ($\xi$). Except for the two OCD$_{GR}$ specific parameters, the settings for the others are discussed by Hinton (2012).

The algorithm first initializes the RBM's free parameters $\Theta$ and the discrete time step $t$ (lines 3–5). Each time step, the algorithm observes a new data instance, collecting $n_B$ new data points into a mini-batch $\mathcal{B}_t$ (lines 8–10). After observing $n_B$ new data points, OCD$_{GR}$ updates the RBM's parameters (line 11–41). The update procedure proceeds in two phases:

**Dynamic generation of historical data (lines 14–22)** As it has been shown in Desjardins et al (2010) and in Cho et al (2010) that RBMs can sample uniformly from the state space, we generate $n_{\hat{B}}$ new training data points to represent past observations based on the data distribution modeled by the RBM at time $t - 1$; these generated data points are collected into the set $\hat{\mathcal{B}}_t$. To obtain good data points with high representational power, we perform Gibbs sampling starting from random values of the hidden neurons drawn from a uniform distribution $\mathcal{U}(0, 1)$.

**CD update with generative replay (lines 26–41)** In the second phase, we update the RBM's weights and biases ($\Theta$) using standard CD for a number of epochs, computing the update only over the most recent mini-batch composed by the union of $\mathcal{B}_t$ and $\hat{\mathcal{B}}_t$. This most recent mini-batch consists of 1.) the data points observed between time $t - 1$ and time $t$ and 2.) the data points generated by the RBM at time $t$. Note that line 39 of Algorithm 1 contains the general form of the update equation, in which $\Psi^+$ (statistics collected from the data) and $\Psi^-$ (statistics collected from the model) can be computed for each free parameter type using Equation 5. Finally, the data points observed between $t - 1$ and $t$, and the data points generated with the RBM at time $t$ are deleted from memory, and OCD$_{GR}$ advances to the next discrete time step $t + 1$ (lines 42–46).

It is easy to observe that an RBM trained with OCD$_{GR}$ acts at any time step $t$ as a generative replay mechanism to provide repetition of approximated past experiences, providing a

```
 1  %% Initialization of the various parameters
 2  Set $n_h, n_v, n_{Gs}, n_{CD}, n_E, n_B, n_{\hat{B}}, \alpha, \rho, \xi$
 3  Initialize RBM parameters $\Theta_0$(i.e., $\mathbf{W}_0, \mathbf{a}_0, \mathbf{b}_0$)$\sim \mathcal{N}(0, \sigma)$
 4  Set $\Delta\Theta_0^{n_E} = 0$
 5  Set $t = 1, \mathbf{B}_t = \varnothing$
 6  %% A continuous loop to handle sequential incoming data
 7  while system is running do
 8  |    Observe a new data point $\mathbf{d}$
 9  |    Add $\mathbf{d}$ to $\mathbf{B}_t$
10  |    if $\mathbf{B}_t$ contains $n_B$ observed data points then
11  |    |    Set $\hat{\mathbf{B}}_t = \varnothing$
12  |    |    %% Generate new data points with the RBM
13  |    |    if $t > 1$ then
14  |    |    |    for $i = 1 : n_{\hat{B}}$ do
15  |    |    |    |    %% Run Gibbs sampling
16  |    |    |    |    Initialize $\mathbf{h} \sim \mathcal{U}(0, 1)$
17  |    |    |    |    for $k = 1 : n_{Gs}$ do
18  |    |    |    |    |    Infer $P(\mathbf{v} = 1|\mathbf{h}, \Theta_{t-1})$
19  |    |    |    |    |    Infer $P(\mathbf{h} = 1|\mathbf{v}, \Theta_{t-1})$
20  |    |    |    |    end
21  |    |    |    |    Add $\mathbf{v}$ to $\hat{\mathbf{B}}_t$
22  |    |    |    end
23  |    |    end
24  |    |    %% Update parameters
25  |    |    Set $\Theta_t^0 = \Theta_{t-1}$ and $\Delta\Theta_t^0 = \Delta\Theta_{t-1}^{n_E}$
26  |    |    for $e(epoch) = 1 : n_E$ do
27  |    |    |    %% Create a training batch from $\mathbf{B}_t$ and $\hat{\mathbf{B}}_t$
28  |    |    |    Set $\mathbf{V} = \mathbf{B}_t \cup \hat{\mathbf{B}}_t$
29  |    |    |    Infer $P(\mathbf{H} = 1|\mathbf{V}, \Theta_t^{e-1})$
30  |    |    |    %% Collect positive statistics $\Psi^+$
31  |    |    |    Compute $\Psi^+$ from $\mathbf{V}$ and $\mathbf{H}$
32  |    |    |    for $k = 1 : n_{n_{CD}}$ do
33  |    |    |    |    Infer $P(\mathbf{V} = 1|\mathbf{H}, \Theta_t^{e-1})$
34  |    |    |    |    Infer $P(\mathbf{H} = 1|\mathbf{V}, \Theta_t^{e-1})$
35  |    |    |    end
36  |    |    |    %% Collect negative statistics $\Psi^-$
37  |    |    |    Compute $\Psi^-$ from $\mathbf{V}$ and $\mathbf{H}$
38  |    |    |    %% Perform parameters update
39  |    |    |    $\Delta\Theta_t^e = \rho\Delta\Theta_t^{e-1} + \alpha[(\Psi^+ - \Psi^-)/(n_B + n_{\hat{B}}) - \xi\mathbf{\Theta}_t^{e-1}]$
40  |    |    |    $\Theta_t^e = \Theta_t^{e-1} + \Delta\Theta_t^e$
41  |    |    end
42  |    |    Set $\Theta_t = \Theta_t^{n_E}$
43  |    |    %% Clean the memory
44  |    |    Delete $\hat{\mathbf{B}}_t, \mathbf{B}_t$ from memory
45  |    |    %% Advance to the next time step
46  |    |    Set $t = t + 1, \mathbf{B}_t = \varnothing$
47  |    end
48  end
```

**Algorithm 1:** Online Contrastive Divergence with Generative Replay. Note that OCD$_{GR}$ only stores the last variant of $\Theta_t^e$ and $\Delta\Theta_t^e$ in memory. Still, we notate them as being indexed by $t$ for a better illustration of the time and training epochs dimensions.

memory-free alternative to ER. In our experiments, we demonstrate empirically that OCD$_{GR}$ can be used successfully to train RBMs in an online setting.

Table 1: Datasets characteristics.

| Dataset | | Dataset Properties | | | |
|---|---|---|---|---|---|
| | | Domain | Features [#] | Train samples [#] | Test samples[#] |
| | MNIST | digits | 784 | 60000 | 10000 |
| | ADULT | households | 123 | 5000 | 26147 |
| | Connect4 | games | 126 | 16000 | 47557 |
| | DNA | biology | 180 | 1400 | 1186 |
| UCI | Mushrooms | biology | 112 | 2000 | 5624 |
| evaluation | NIPS-0-12 | documents | 500 | 400 | 1240 |
| suite | OCR-letters | letters | 128 | 32152 | 10000 |
| | RCV1 | documents | 150 | 40000 | 150000 |
| | Web | Internet | 300 | 14000 | 32561 |

### 3.3 Computational Complexity

The primary difference between ER and OCD$_{GR}$ at each discrete time step $t$ is that ER has to recall random data from memory, while OCD$_{GR}$ generates the data via Gibbs sampling. For ER, the memory recall time depends upon the hardware platform and the programming environment (e.g., Matlab, C++), and so is not easily quantifiable. For OCD$_{GR}$, the *dynamic generation of historical data* phase using Gibbs sampling requires, on one side, a small number of matrix multiplication (which may be parallelized) and are linearly dependent by $n_{Gs}$ and, on the other side, the computation of the sigmoid functions for the visible and hidden neurons. This yields a per-update time of $\mathcal{O}(2n_{Gs}n_v n_h + n_{Gs}n_h + n_{Gs}n_v)$, which in the typical case of $n_{Gs} = 1$, reduces to $\mathcal{O}(n_v n_h)$.

## 4 Experiments and Results

Firstly, we considered a toy scenario (i.e., an artificially generated dataset) to illustrate the OCD$_{GR}$ behavior. Secondly, we evaluated OCD$_{GR}$ performance on the MNIST dataset[1] of handwritten digits, and on the UCI evaluation suite (Germain et al, 2015). Thus, overall, the evaluation was performed on 9 datasets coming from different domains, which are detailed in Table 1.

To simulate the online learning setting, each training instance was fed to the RBM training algorithm only once in a sequential manner in one of two orders: 1.) a worst-case scenario, in which the data instances are presented in order of the classes, and 2.) a more realistic scenario, in which the instances are ordered randomly.

The update procedure of the RBM's free parameters was triggered each time after the system had observed and collected 100 data points (i.e., $n_B = 100$). To find the best meta-parameters specific for OCD$_{GR}$ (i.e., $n_{\hat{B}}$, $n_{Gs}$) we conducted a random search. Based on this small experiment, before each update procedure took place, we generated another $n_{\hat{B}} = 300$ data points according with Algorithm 1, lines 14–22, with $n_{Gs}$ set to 1. Moreover, another reason to set a small number of steps for Gibbs sampling when new data points are generated (i.e., $n_{Gs} = 1$) is given by the fact that if we use samples from the model for both components of the gradient (i.e., Equation 3), these will cancel out in expectation. Except when specified otherwise, the other meta-parameters used usually in the RBM training process were set to standard values, such as $n_E = 10$, $n_{CD} = 1$, $\alpha = 0.05$, $\rho = 0.9$ (except the first 5 training epochs in which $\rho = 0.5$), and $\xi = 0.0002$, following (Hinton, 2012). Please note that even a higher number of contrastive divergence steps (i.e., $n_{CD}$) may lead to a better performance on some specific datasets, i.e., MNIST, it leads also to an increasing amount of computations. As our goal was to propose a fast algorithm to train RBMs in an online manner also on low-resources devices, we preferred to perform most of our experiments using just 1 step contrastive divergence.

---
[1] http://yann.lecun.com/exdb/mnist/. Last visit on 26 September 2016.

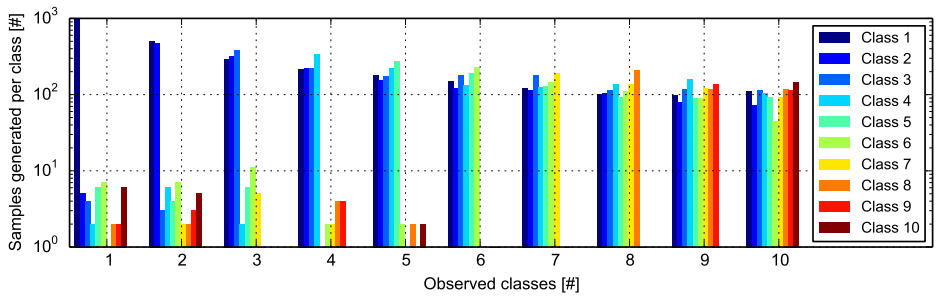

Fig. 3: Illustration of the OCD$_{\text{GR}}$'s behavior (toy scenario). At any time, the samples generated by $RBM_{OCD}$ are distributed equally among all observed classes. The y-axis uses the log-scale.

### 4.1 Illustration of OCD$_{\text{GR}}$'s behavior (toy scenario)

To easy visualize the quality of the samples generated by RBMs trained with OCD$_{\text{GR}}$ ($RBM_{OCD}$) we have considered a toy scenario with artificially generated data and an RBM with 100 visible neurons and 50 hidden neurons. For training we have created 10000 data points (each data point being a binary vector of 100 elements) split in 10 classes of 1000 data points each, as following. For Class 1 $p(v_i = 1) = 0.3 \Leftrightarrow 1 \leq i \leq 10$ and $p(v_i = 1) = 0 \Leftrightarrow 11 \leq i \leq 100$, and so on up to Class 10 for which $p(v_i = 1) = 0.3 \Leftrightarrow 91 \leq i \leq 100$ and $p(v_i = 1) = 0 \Leftrightarrow 1 \leq i \leq 90$. During training, we have firstly observed all data instances belonging to Class 1, and after that we have generated 1000 samples with the trained $RBM_{OCD}$. Next, we have continued the training procedure using all data points belonging to Class 2, and then we have generated another 1000 samples, and further on we repeated this procedure until all 10 classes have been considered. To classify the samples generated by $RBM_{OCD}$ we used k-nearest neighbors. Figure 3 shows that OCD$_{\text{GR}}$ behaves as expected and as new classes are observed the $RBM_{OCD}$ enlarges its encoded distribution.

### 4.2 Evaluation

We compared our proposed method, $RBM_{OCD}$, against 1.) RBMs trained using Experience Replay with a Memory Limit ($RBM_{ER\text{-}ML}$) and 2.) RBMs trained using Experience Replay with Infinite Memory ($RBM_{ER\text{-}IM}$). For a fair comparison, in the case of $RBM_{ER\text{-}ML}$ we limited the number of data points stored in memory to occupy approximately the same number of bytes as the parameters of $RBM_{OCD}$. We highlight that by allowing $RBM_{ER\text{-}ML}$ to have an experiences memory of the same size with the $RBM_{OCD}$ parameters means that, in fact, $RBM_{ER\text{-}ML}$ needs a double memory size than $RBM_{OCD}$, as it needs also some memory to store its own parameters. In contrast, we allow $RBM_{ER\text{-}IM}$ to store all observed data points

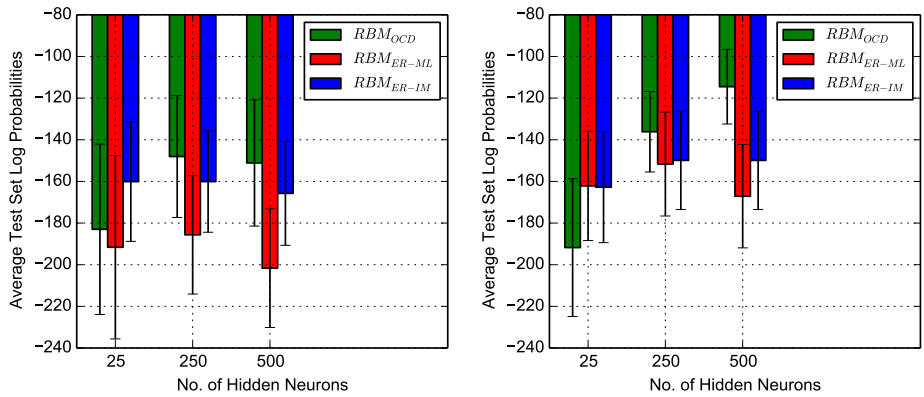

(a) Worst case scenario, with instances in ascending order by digit.

(b) Realistic scenario, where the instances are ordered randomly.

Fig. 4: Performance on the MNIST dataset. For each model, we plot the average log probabilities computed on the entire test set, with error bars representing the standard deviation of the average log probabilities computed on each digit class.

in memory. To train both experience replay models, i.e., $RBM_{ER\text{-}ML}$ and $RBM_{ER\text{-}IM}$, we use a similar algorithm to Algorithm 1. The only main difference is that instead of generating new samples using the RBM models themselves (Algorithm 1, lines 11-23), we retrieve those samples from the replay memory. Moreover, same as for $RBM_{OCD}$, in the case of $RBM_{ER\text{-}ML}$ and $RBM_{ER\text{-}IM}$ models, we used 300 randomly chosen data points from the memory of past experiences and the same meta-parameters values. To quantify the generative performance of the trained networks, we used Annealed Importance Sampling (AIS) with the same parameters as in the original paper (Salakhutdinov and Murray, 2008) to estimate the partition function of the RBMs and to calculate their log probabilities. On each dataset, after all training data points were given to the learner, we computed the average log probabilities on the entire test set.

*4.2.1 Worst Case Scenario: Sorted Order*

In the first scenario, we have used the binarized MNIST dataset. During training, the data instances were ordered sequentially in ascending order of the digits $(0, 1, \ldots, 9)$, making it a difficult scenario for online learning. For each algorithm, we considered various numbers of hidden neurons ($n_h \in \{25, 250, 500\}$), and 784 visible neurons (i.e., $28 \times 28$ binary image). Figure 4a shows that $RBM_{OCD}$ outperforms $RBM_{ER\text{-}ML}$ in all cases, independent of the number of hidden neurons. Moreover, it outperforms even $RBM_{ER\text{-}IM}$ when it has enough representational power (i.e., 250 and 500 hidden neurons). It is interesting to see that while the generative power of $RBM_{OCD}$ increases with the number of hidden neurons, $RBM_{ER\text{-}IM}$ is not significantly affected when given more hidden neurons. Further, the $RBM_{ER\text{-}ML}$ model *loses* its generative power when the number of hidden neurons is increased. These results may be explained by the fact that having more hidden neurons helps $RBM_{OCD}$ to better model the data distribution. In contrast, in the case of experience replay mechanisms with memory, a larger RBM would need more past-experience training data to avoid forgetting the distribution of the first observed data points, especially in the case of $RBM_{ER\text{-}ML}$. This situation does not occur for $RBM_{OCD}$, due to the fact the data points generated randomly by the RBM itself using Gibbs sampling approximate well the distribution of the past-experience data. For the sake of clarification, we mention that even if at a first look an inter-class standard deviation of $30-40$ *nats* for all online trained models is striking, in fact, it is same as the one obtained for the offline trained RBMs.

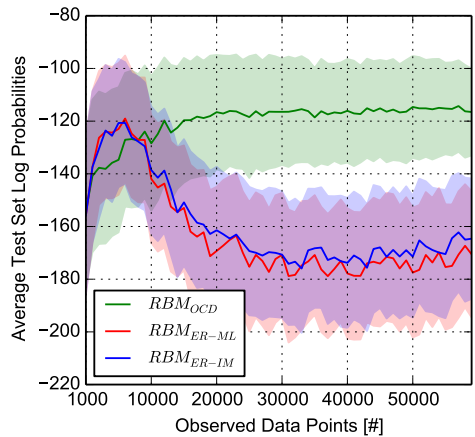

Fig. 5: Model performance over time on the realistic scenario using the MNIST dataset. The straight lines represent the average log probabilities computed on the entire test set, while the shadowed areas represent the standard deviation of the average log probabilities computed on each digit class.

*4.2.2 Realistic Scenario: Random Order*

In the second more realistic scenario, the training instances were presented sequentially in random order. As this is an usually encountered situation, herein, besides the MNIST dataset, we have used also the UCI evaluation suite (Germain et al, 2015). The latter one contains contains 8 real-world binary datasets from various domains, specially selected to evaluate the performance of density estimation models.

**MNIST dataset.** Figure 4b shows that $RBM_{OCD}$ outperforms both $RBM_{ER\text{-}ML}$ and $RBM_{ER\text{-}IM}$, when it has enough representational power (i.e., 250 and 500 hidden neurons). As in the previous experiment, the generative performance of $RBM_{OCD}$ increases as the number of hidden neurons increases, but the gain in performance is even higher in this situation, culminating with $-114.53$ *nats* in the case of an RBM with 500 hidden neurons. Thus, $RBM_{OCD}$ outperforms $RBM_{ER\text{-}IM}$ by $35.39$ *nats* at the same number of hidden neurons. In fact, $RBM_{OCD}$ with 500 hidden neurons outperforms by $11.01$ *nats* even the state-of-the-art results reported by Salakhutdinov and Murray (Salakhutdinov and Murray, 2008) (see Table 2, third row) for an RBM with 500 hidden neurons trained completely offline with standard one-step contrastive divergence. Besides the improved average performance on the entire test set, also observe that as the number of hidden neurons increases in the case of $RBM_{OCD}$, the standard deviation (computed on the average log probabilities from each digit class) decreases. The smaller standard deviation implies that the model represents all classes well, without imbalance.

Table 2: Realistic scenario. The results are given for RBMs with $n_h = 500$ and $n_{CD} = 1$. On the MNIST dataset the offline RBM results are taken from Salakhutdinov and Murray (2008), while on the UCI evaluation suite the offline RBMs results are taken from Germain et al (2015).

|  | Dataset | Online models | | | Offline model |
| --- | --- | --- | --- | --- | --- |
|  |  | $RBM_{OCD}$ | $RBM_{ER\text{-}IM}$ | $RBM_{ER\text{-}ML}$ | $RBM$ |
|  | MNIST | **-114.52** | -151.67 | -167.11 | -125.53 |
|  | ADULT | -19.64 | -18.08 | **-17.28** | -16.26 |
|  | Connect4 | -16.28 | **-16.03** | -17.64 | -22.66 |
|  | DNA | **-103.14** | -111.81 | -114.84 | -96.74 |
| UCI | Mushrooms | **-16.64** | -20.38 | -17.58 | -15.15 |
| evaluation | NIPS-0-12 | **-290.06** | -365.03 | -339.82 | -277.37 |
| suite | OCR-letters | **-47.61** | -51.35 | -53.85 | -43.05 |
|  | RCV1 | **-53.28** | -56.34 | -79.06 | -48.88 |
|  | Web | -33.47 | **-32.58** | -35.07 | -29.38 |

To better understand $RBM_{OCD}$'s behavior, we performed an additional experiment in the realistic scenario. We again trained $RBM_{OCD}$, $RBM_{ER\text{-}ML}$, and $RBM_{ER\text{-}IM}$ models, each with 500 hidden neurons. However, in this experiment, we measured performance on the MNIST test set during the training phase after every 1,000 observed data points. Figure 5 shows an interesting behavior for all three models. The $RBM_{OCD}$ has a very stable learning curve which increases over time. In contrast, $RBM_{ER\text{-}ML}$ and $RBM_{ER\text{-}IM}$ show unstable learning curves. This behavior can be explained by the fact that when the probability of selecting for replay any past observed data point decreases below a certain threshold, then the subset of the selected data points for replay no longer represents well the distribution of the past-experience data, and the models become over-fitted. To avoid this situation, the number of selected data points from the replay memory would need to increase linearly with the number of observations. However, this solution is infeasible as it will lead to a linear increase in the computational complexity of $RBM_{ER\text{-}ML}$ and $RBM_{ER\text{-}IM}$ over time, leading to non-realistic online learning algorithms. In contrast, $RBM_{OCD}$ is a Markov chain and it is not affected by this situation, explaining why $RBM_{OCD}$ outperforms $RBM_{ER\text{-}ML}$ and $RBM_{ER\text{-}IM}$ after observing approximately 8,000 instances.

In our final experiment on the MNIST dataset, we varied the number of contrastive divergence steps, training an $RBM_{OCD}$ with 500 hidden neurons using 3 steps and 10 steps of contrastive divergence. Similarly to the RBM's behavior reported by Salakhutdinov and Murray (Salakhutdinov and Murray, 2008), further CD steps improved the generative performance of these models. In the case when $n_{CD} = 3$, the average log probabilities on the MNIST test set was -108.96; for $n_{CD} = 10$, it was -104.31.

**UCI evaluation suite**. Herein, we have trained $RBM_{OCD}$, $RBM_{ER\text{-}ML}$, and $RBM_{ER\text{-}IM}$ with $n_h = 500$, and $n_v$ set to the number of features of each dataset. The results reflected in Table 2 show that $RBM_{OCD}$ outperforms $RBM_{ER\text{-}ML}$ and $RBM_{ER\text{-}IM}$ on 5 out of 8 datasets, while on the other 3 has a very close generative performance to the top performer. Overall, we may observe that as the size of the dataset increases, or as data distribution became more complex, $RBM_{OCD}$ starts having a clear advantage over $RBM_{ER\text{-}ML}$ or $RBM_{ER\text{-}IM}$.

In all experiments performed, we observed that our MATLAB implementations of all algorithms ran in approximately the same time. Given the same RBM configuration, $RBM_{ER\text{-}IM}$ was slightly slower then $RBM_{OCD}$, which was slightly slower then $RBM_{ER\text{-}ML}$. However, the differences were on the order of few milliseconds, with one note that for $RBM_{ER\text{-}IM}$, the difference increased with the number of data points saved in memory.

## 5 Conclusion

We have proposed a novel method, Online Contrastive Divergence with Generative Replay (OCD$_{GR}$), to train RBMs in online settings. Unlike current experience replay mechanisms which directly recall recorded observations from memory, OCD$_{GR}$ uses the generative capabilities of RBMs to dynamically simulate past experiences. As a consequence, it does not need to store past observations in memory, substantially reducing its memory requirements. We demonstrated that RBMs trained online with OCD$_{GR}$ outperform RBMs trained online using experience replay mechanisms with memory, with few exceptions, in which their performance is comparable. We highlight that in some cases they even outperform RBMs having a similar number of hidden neurons, but trained *offline* with standard contrastive divergence.

In future work, we intend to understand better the effect of the various OCD$_{GR}$ metaparameters (especially the relation between the number of generated samples and the number of observed samples) on the RBM's generative performance, and to extend OCD$_{GR}$ to other suitable generative models, e.g., deep Boltzmann machines, autoencoders. Other interesting research directions, which we hope to investigate, is to use to use RBMs trained with OCD$_{GR}$ to perform anomaly detection in low-resources devices, to control DRL algorithms by generating RL atomic operations instead of using experience replay mechanisms with memory to store them, to perform online supervised learning by generating input-output pairs for the online training of deep networks.

**Acknowledgements** This research has been partly funded by the European Union's Horizon 2020 project INTER-IoT (grant number 687283). A portion of this work has taken place in the Learning Agents Research Group (LARG) at UT Austin. LARG research is supported in part by NSF (CNS-1330072, CNS-1305287, IIS-1637736, IIS-1651089), ONR (21C184-01), and AFOSR (FA9550-14-1-0087).